\documentclass[runningheads]{ola}

\usepackage{epsfig}
\usepackage{graphicx} 
\usepackage{algorithm}
\usepackage[utf8]{inputenc}
\usepackage{algpseudocode}
\usepackage{color,soul}

\begin{document}

\pagestyle{headings}

\mainmatter

%\title{Learning to compute inner consensus \\ A novel approach to modeling agreement between Capsules\thanks{Supported by Calouste Gulbenkian Foundation}}

% Title
\titlerunning{Learning to compute inner consensus}

% Title for odd pages
\author{Gonçalo Faria \\ \email{goncalorafaria@tecnico.ulisboa.pt}}
%\author{Gonçalo Faria \\ \email{goncalorafaria@gmail.com}}

\title{Learning to compute inner consensus \\ A novel approach to modeling agreement between Capsules\thanks{Supported by Calouste Gulbenkian Foundation}}
\date{}

% Authors for the top of the even pages
\authorrunning{G. Faria}

\institute{Instituto Superior Técnico\\ Lisboa, Portugal }
%\institute{Department of Informatics\\ School of Engineering \\ University of Minho }

\maketitle

\begin{abstract}

    This project considers Capsule Networks, a recently introduced machine learning model that has shown promising results regarding generalization and preservation of spatial information with few parameters. 
    The Capsule Network's inner routing procedures thus far proposed, a priori, establish how the routing relations are modeled, which limits the expressiveness of the underlying model. 
    In this project, I propose two distinct ways in which the routing procedure can be learned like any other network parameter. 

\end{abstract}

\section{Introduction}

Starting with the developments made by Frank Rosenblatt surrounding the Perceptron algorithm \cite{Rosenblatt1958}, 
innovative techniques have marked the beginning of a new, biologically inspired, approach in Artificial Intelligence that is, surprisingly, better suited to deal with naturally unstructured data.

The now called field of Deep Learning has expanded these ideas by creating models that stack multiple layers of Perceptrons. These Multilayer Perceptrons, commonly known as Neural Networks, achieve greater representation capacity, due to the layered manner the computational complexity is added, especially when compared with its precursor. Attributable to this compositional approach they are especially hard-wired to learn a nested hierarchy of concepts. 

Aided by the increase in computational power as well as efforts to collect high quality labeled data, in the last decade, a particular subset of Neural Networks, called Convolutional Neural Networks(\textbf{CNN}) \cite{LeCun1989}, have accomplished remarkable results \cite{Krizhevsky2012}. As their common trait, having to deal with high dimension unstructured data, computer vision \cite{Krizhevsky2012}, speech recognition \cite{HintonG.DengL.YuD.DahlG.MohamedA.JaitlyN.Kingsbury2012}, natural language processing \cite{Bengio2003}, machine translation \cite{Google} and medical image analysis \cite{Kleesiek2016,Dou2016} are the fields in which these models have shown greater applicability.

Colloquially, a CNN as presented by Yann LeCun and others, is a model that uses multiple layers of feature detectors that have local receptive fields and shared parameters interleaved with sub-sampling layers \cite{Krizhevsky2012,Sermanet2011,Szegedy2015}. 
For attaining translation invariance, by design, these sub-sampling layers discard spatial information \cite{Hinton2011}, which, when applied to the classification task, assist in amplifying the aspects of its input that are useful for discriminating and suppress irrelevant variations that are not.

Translation invariance, 
however helpful in attaining a model that has the same classification when applied to entities in different viewpoints, 
inevitably requires training on lots of redundant data. 
This redundancy is artificially introduced to force the optimization process to find solutions that can not distinguish between different viewpoints of the same entity. 
Additionally, disregarding spatial information produces models incapable of dealing with recognition tasks, 
such as facial identity recognition, that require knowledge of the precise spatial relationships between high-level parts, like a nose or a mouth. 

To address these drawbacks, adaptations on CNNs  have been proposed. 
This project focuses on improving an existing equivariant approach, 
introduced by Sara Sabour, Geoffrey E. Hinton and Nicholas Frosst, 
called Capsule Networks \cite{Sabour2017,Hinton2018}.

The Capsule Network model proposed by Sabour et al. \cite{Sabour2017} uses, layer wise, dynamic routing. 
Since dynamic routing is a sequential extremely computationally expensive procedure, 
if the network were to be scaled, in order to be suited to solve more challenging datasets, 
would incur in overly expensive costs in training and inference time. 
Furthermore, when applied to many capsules, 
the gradient flow through the dynamic routing computations is dampened. 
This inhibits learning, regardless of the computational resources used. 
Additionally, both dynamic routing as well as other routing procedures proposed, 
a priori, establish the way in which the routing relations are modeled, which limits the expressiveness of the underlying model. 

In this work, I propose two distinct ways in which the routing procedure can be discriminatively learned. 
In parallel with \cite{Hinton2018}, 
I employ routing to local receptive fields with parameter sharing, 
in order to reduce vanishing gradients, 
take advantage of the fact that correlated capsules tend to concentrate in local regions and reduce the number of model parameters.

\section{Related Work}

Capsules were first introduced in \cite{Hinton2011}, whereas the logic of encoding instantiation parameters was
established in a transforming autoencoder.

More recently, further work on capsules \cite{Sabour2017} garnered
some attention achieving state-of-the-art performance on MNIST, with a shallow Capsule Network
using an algorithm named Dynamic routing.

Shortly thereafter, a new routing algorithm, based on an Expectation-Maximization extension applied to a Gaussian Mixing Model, was proposed in \cite{Hinton2018}.
Additionally, capsule vectors were replaced by matrices to reduce the number of parameters and also convolutional capsules were introduced. 
State-of-the-art performance was achieved on the smallNORB dataset using a relatively small Capsule Network.

An analysis of Dynamic routing algorithm as an optimization problem was presented in \cite{Wang2018} as  well as a discussion of possible ways to improve Capsule Networks.
Furthermore, researchers have proposed several extentions \cite{av1,av2,av3,av4,av5,av6,av7,av8} showing that there is still room for improvement.

\section{ Capsule Networks }

In parallel with a neural network, a Capsule Network \cite{Sabour2017,Hinton2018} is, in essence, multiple levels of capsule layers which, in part, are composed of various capsules. 
A capsule is a group of artificial neurons which learns to recognise an implicitly defined entity, in the network’s input, and outputs a tensor representation, the \textbf{pose}, which captures the properties of that entity relative to an implicitly defined canonical version and an activation probability. The activation probability was designed to express the presence of the entity the capsule represents in the network’s input. In the high-level capsules this activation probability corresponds to the inter-class invariant discriminator used for classification.
Moreover, every capsule’s pose is \textbf{equivariant}, meaning that as the entity moves over the appearance manifold, the pose moves by a corresponding amount.  

Every capsule layer is either calculated from multiple feature maps, when they correspond to the \textbf{primary capsule layer}, or by a series of transformations, followed by a mechanism called \textbf{routing}, applied to the outputs of the previous capsule layer.

Disentangling the internal representations in viewpoint invariant, presence probability and viewpoint equivariant, instantiation parameters may prove to be a more generalizable approach for representing knowledge than the conventional CNN view, that only strives for representational invariance. Early evidence of this has been revealed by experimenting on the effects of varying individual components of the individual capsules. The MNIST trained network presented in \cite{Sabour2017}, without being explicitly designed to, learned to capture properties such as stroke, width and skew.  In addition to this, as mentioned  by Hinton and others \cite{Sabour2017}, the same network, despite never being trained with digits, that were subject to affine transformations, was able to accurately classify them during test time, which seems to suggest greater generalization capacity when in comparison with conventional CNNs. 

\subsection{Routing Procedure}
\label{routing_section}

Routing consists of a dot-product self-attention \cite{Vaswani2017AttentionIA} procedure that assigns, for each output capsule, a distribution of \textbf{compatibility probabilities} to the transformed previous layer's capsules, the \textbf{capsule votes}. These compatibility probabilities, after multiplied to corresponding capsule votes are combined and result in the output capsule's pose. Furthermore, the routing procedure also assigns the activation probability to the respective output capsule, usually based on the amount of agreement between the votes with higher compatibility probabilities.

\begin{figure}
    \centering  
\includegraphics[scale=0.18]{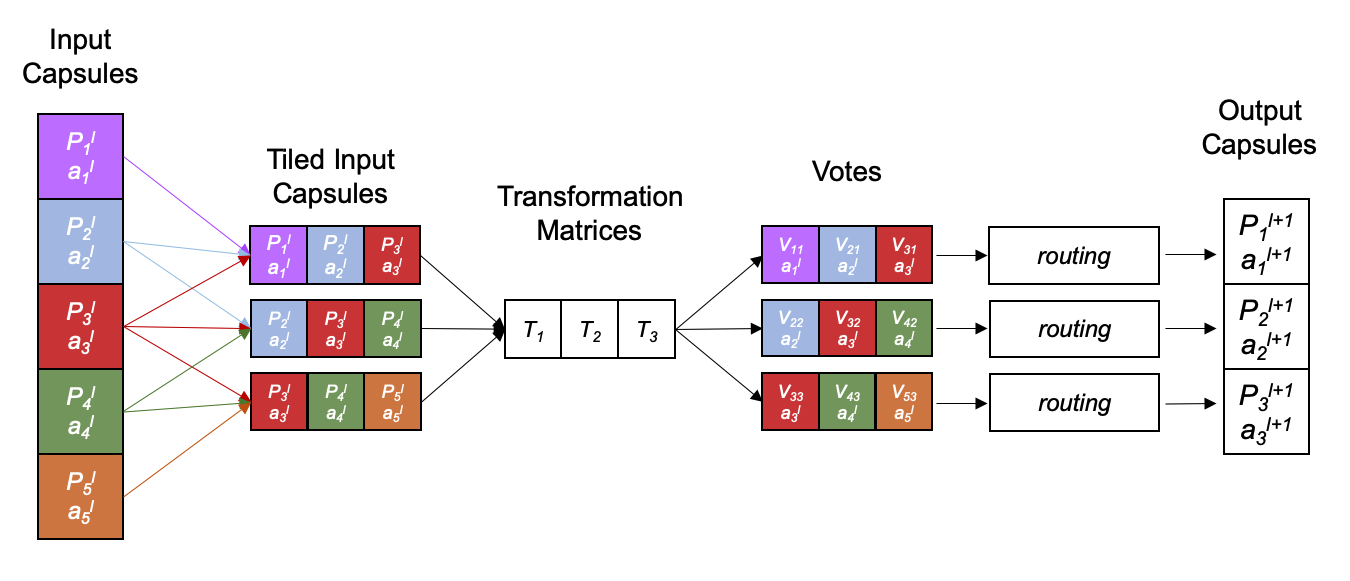}
\caption{1D Convolutional Routing(adapted from \cite{ibm2019}). Respectively, $P_{i}^{l}$ and $a_{i}^{l}$, denote the pose and activation corresponding to the $ith$ capsule in the $l$ capsule layer.}
\label{fig:1drout}
\end{figure}

This procedure provides the framework for a consensus mechanism in the multi-layer capsule hierarchy, which, for the higher levels of the input’s domain, will solve the problem of assigning parts to wholes.
Additionally, it can be applied to local receptive fields with shared transformation matrices. Figure \ref{fig:1drout} contains a diagram representing how routing is applied convolutionally in one dimension. The 2D and 3D convolutional routing is extrapolated in the same manner as the usual 2D and 3D convolution would.

The vote transformations are traditionally linear transformations. 
However, due to some stability issues that occur during training, equation \ref{eq:transf}, proposed by \cite{Wang2018}, will be used instead. The capsule vote from the $ith$ input capsule to the $jth$ output capsule is obtained by multiplying the matrix $W_{ij}$ to the pose of the $ith$ input capsule $P_i$ divided by the frobenius norm of $W_{ij}$.
 
\begin{equation}\label{eq:transf}
    v_{ij} =  \frac{ W_{ij} }{ \| W_{ij} \|_\mathcal{F}} P_i
\end{equation}

In algorithm \ref{agreementgen}, I present a generalization of the routing procedure that encompasses most of the routing algorithms proposed. The definition of the $activation$ and $compatibility$ functions is what caracterizes the particular routing procedure.

\begin{algorithm}{
    The routing algorithm returns the output capsule's pose $\mu$ and activation $p$ given a subset of capsules $\gamma$. 
    For some capsule $i \in \gamma$ the capsule activation is represented
    by $a_i$, the capsule vote is represented by $v_i$ and 
    the compatibility probability is represented by $c_i$. 
    The set of all $v_i$, $a_i$ and $c_i$ is represented, respectively, by $V$, $A$, $C$.  
    The symbol $S^t$ represents some state values at timestep $t$
    that might be shared either by the compatibility function across iterations or by the compatibility and activation functions. 
}
\caption{Generic Routing Mechanism} \label{agreementgen}
\begin{algorithmic}[1]
        \Procedure{Routing}{$V, A, \gamma$}
            \State $\forall i \in \gamma : c_{i} \gets 1 / \| \gamma \|$ 
            \State $s^0 \gets s_0$
            \State $ \mu \gets \frac{\sum_{ i \in \gamma }^{} c_{i} v_{i}}{ \sum{ i \in \gamma }^{} c_{i} }$
            \State $ p \gets activation_\beta(\mu, s, V, c, \gamma)$
            \For{$ t$ iterations }
                \State $\forall i \in \gamma : c_{i}, s_{i}^{t} \gets compatibility_\theta(c_{i}, s_{i}^{t-1}, v_{i}, a_i, \mu, p)$
                \State $ \mu \gets \frac{\sum_{ i \in \gamma }^{} c_{i} v_{i}}{ \sum{ i \in \gamma }^{} c_{i} }$
                \State $ p \gets activation_\beta(\mu, S^{t}, V, C, \gamma, A)$
            \EndFor
            \State \Return $\mu, p$
        \EndProcedure

    \end{algorithmic}
\end{algorithm}

\section{Learning the Routing Procedure}
In hopes of improving the performance of Capsule Networks and of incorporating routing into the whole training process, instead of designing an alternative routing procedure, 
that necessarily constrains the parts to whole relationships that can be modeled, 
I present methods for parameterizing it, such that, 
either for each layer or each network, the routing procedure itself can be discriminatively learned like any other model parameter. 

In the following subsections I present two distinct alternatives. 
The first one exposes routing as a classic clustering algorithm based on the application of parametric kernel functions.
The second takes a less structured approach that defines the activation and compatibility functions simply as neural networks.

\subsection{Similarity Learning Approach}
\label{klearning}

Since for each input, in every output capsule, 
the routing computations are reminiscent of an agglomerative fuzzy clustering algorithm, 
following the avenue taken in \cite{Wang2018}, in this subsection, routing it’s analyzed as the optimization of a clustering-like objective function. The resulting cluster is interpreted as the agreement over the capsule votes and is used as the routing procedure's output capsule's pose.
Similarly with \cite{Wang2018} I propose \ref{eq:loss}, inspired by the algorithm presented in \cite{Li2008}.

\begin{equation} \label{eq:loss}
    \min \mathcal{L}(C,\mu) = - \sum_{i=1}^{n} c_i \langle \mu , v_i \rangle_{\theta} + \lambda_1 D_{KL}(C\parallel U) + \lambda_2 D_{KL}(C\parallel A)
\end{equation}

subject to

\begin{equation} \label{eq:const}
\sum_{i=1}^{n} c_i = 1, c_i \in [0,1 ], 1 \leq i \leq n
\end{equation}
where $\langle \cdot , \cdot \rangle_{\theta}$ is a kernel function defined by $\theta$, $\mu$ is the output capsule's pose, $v_i$, $a_i$ and $c_i$ are, respectively, 
the $ith$ input capsule vote, its activation and compatibility probability. $A = [a_i]$ and $C = [c_i]$ are vectors of \textbf{n} components. 
The compatibility probability corresponds to the weight of the $ith$ input capsule vote has in 
the output capsule's pose. The parameters $\lambda_1$ and $\lambda_2$, which are both positive or equal to zero, are the weights of the penalty terms applied to $C$. The symbol $U$ denotes the uniform distribution.

The first term in the objective function is the weighted average of the similarity between the output pose and the $ith$ input votes. 
The weights are the compatibility probabilities. 
In this way, the more compatible an input capsule is the more significant its vote similarity is in the optimization process. 
The remaining terms are penalty terms weighted by $\lambda_1$ and $\lambda_2$. The second term, as in \cite{Li2008}, 
is proportional to the Kullback–Leibler(KL) divergence\cite{kldv} of the compatibility probabilities 
and the uniform distribution. Additionally, the third term is proportional to the KL diverge between the compatibility probabilities and 
the activations. In other words, the expectation of the logarithmic difference between the $A$ and $C$, where the expectation 
is taken with respect to $C$.
The penalty terms were introduced in order to pay, independently, both for the nonalignment between the 
activations and compatibility probabilities aswell as the nonuniformity of the compatibility probabilities.

The minimization problem is solved by partially optimizing $\mathcal{L}(C,\mu)$ for $C$ and $\mu$.

    For a fixed $C$, $\mu$ is updated as 
    
\begin{equation}
    \mu = \sum_{i=1}^{n} c_i v_i
\end{equation}

For a fixed $\mu$, $C$ is updated as follows: Using the Lagrangian multiplier technique, 
We obtain the unconstrained minimization problem \ref{eq:lagrangean}.

\begin{equation} \label{eq:lagrangean}
\tilde{\mathcal{L}}(C,\alpha) = - \sum_{i=1}^{n} c_i \langle \mu , v_i \rangle_{\theta} +  \lambda_1 D_{KL}(C\parallel U) + \lambda_2 D_{KL}(C\parallel A) - \alpha (\sum_{i=1}^{n} c_i - 1), \lambda_2 >= 0, \lambda_1 >= 0
\end{equation}
where $\alpha$ is the Lagrangian multiplier. 
If $(\hat{C},\hat{\alpha})$ is a minimizer of $\tilde{\mathcal{L}}(C,\alpha)$ then the gradient must be zero. Thus,
    
 \begin{equation} \label{eq:gradc}
    \frac{\partial \tilde{\mathcal{L}}(\hat{C},\hat{\alpha})}{ \partial c_i} = - \langle \mu , v_i \rangle_{\theta} + (\lambda_1 +\lambda_2 )( 1 + log \hat{ c_i}) - \lambda_2 log a_i - \hat{\alpha} = 0, 1\leq i \leq n \\
\end{equation}   

and

\begin{equation} \label{eq:gradl}
    \frac{\partial \tilde{\mathcal{L}}(\hat{C},\hat{\alpha})}{ \partial \alpha} = \sum_{i=1}^{n} \hat{ c_i } - 1 = 0
\end{equation}

From \ref{eq:gradc}, we obtain

\begin{equation} \label{eq:expl}
    \hat{c_i} =  exp( \frac{\langle \mu , v_i \rangle_{\theta}}{\lambda_1+\lambda_2}) exp( \frac{ \hat{ \alpha}}{\lambda_1+\lambda_2} ) a_i^{\frac{\lambda_2}{\lambda_1+\lambda_2}}  exp(-1), 1 \leq i \leq n
\end{equation}

By substituting \ref{eq:expl} into \ref{eq:gradl}, we have
    
\begin{equation}
    \sum_{i=1}^{n} \hat{ c_i } =exp(-1) exp( \frac{ \hat{ \alpha}}{\lambda_1+\lambda_2} ) \sum_{i=1}^{n} exp( \frac{\langle \mu , v_i \rangle_{\theta}}{\lambda_1+\lambda_2}) a_i^{\frac{\lambda_2}{\lambda_1+\lambda_2}} = 1
\end{equation}
which, when solved for $\hat{\alpha}$, results in

\begin{equation} \label{eq:alpha}
    \hat{ \alpha} = (\lambda_1+\lambda_2)log \frac{1}{ exp(-1) \sum_{j = 1}^{} exp(\frac{\langle \mu , v_j \rangle_{\theta}}{\lambda_1+\lambda_2} ) a_j^{\frac{\lambda_2}{\lambda_1+\lambda_2}} } 
\end{equation}

It follows that by substituting \ref{eq:alpha} into \ref{eq:expl}, leading to

\begin{equation} \label{eq:update}
    \hat{c_i} = \frac{ a_i^{\frac{\lambda_2}{\lambda_1+\lambda_2}} exp( \frac{\langle \mu , v_i \rangle_{\theta}}{\lambda_1+\lambda_2}) }{ \sum_{j=1}^{n} a_j^{\frac{\lambda_2}{\lambda_1+\lambda_2}} exp( \frac{\langle \mu , v_j \rangle_{\theta}}{\lambda_1+\lambda_2}) }, 1 \leq i \leq n
\end{equation}
and $C$ can be updated by \ref{eq:update}.

\begin{figure}
    \centering
        \begin{tabular}{c c}
            (a) Optimal Values of C for different values of $\lambda_1$ and $\lambda_2$. & (b) Distribution of activations. \\
            \includegraphics[scale=0.25]{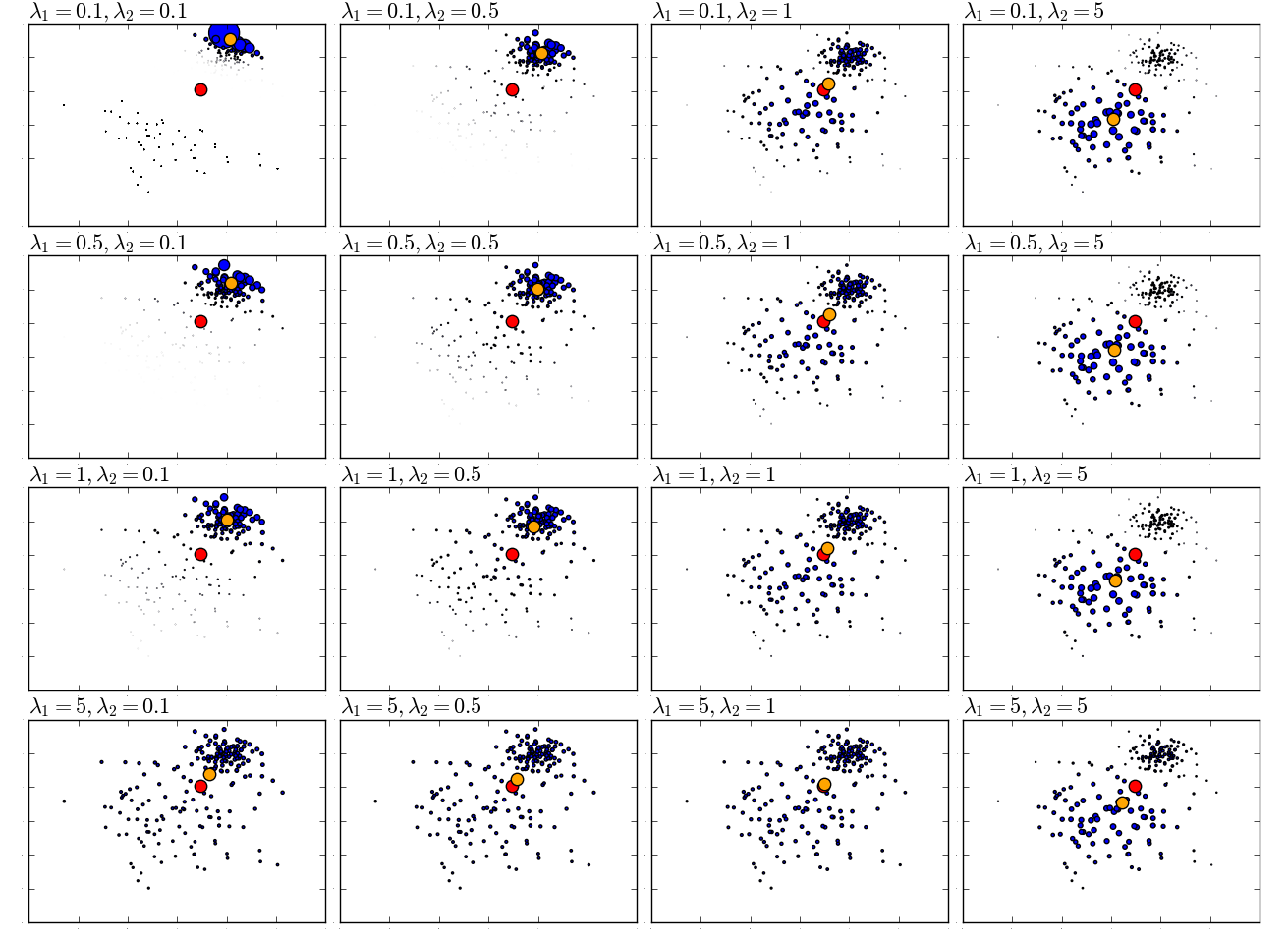} & \includegraphics[scale=0.4]{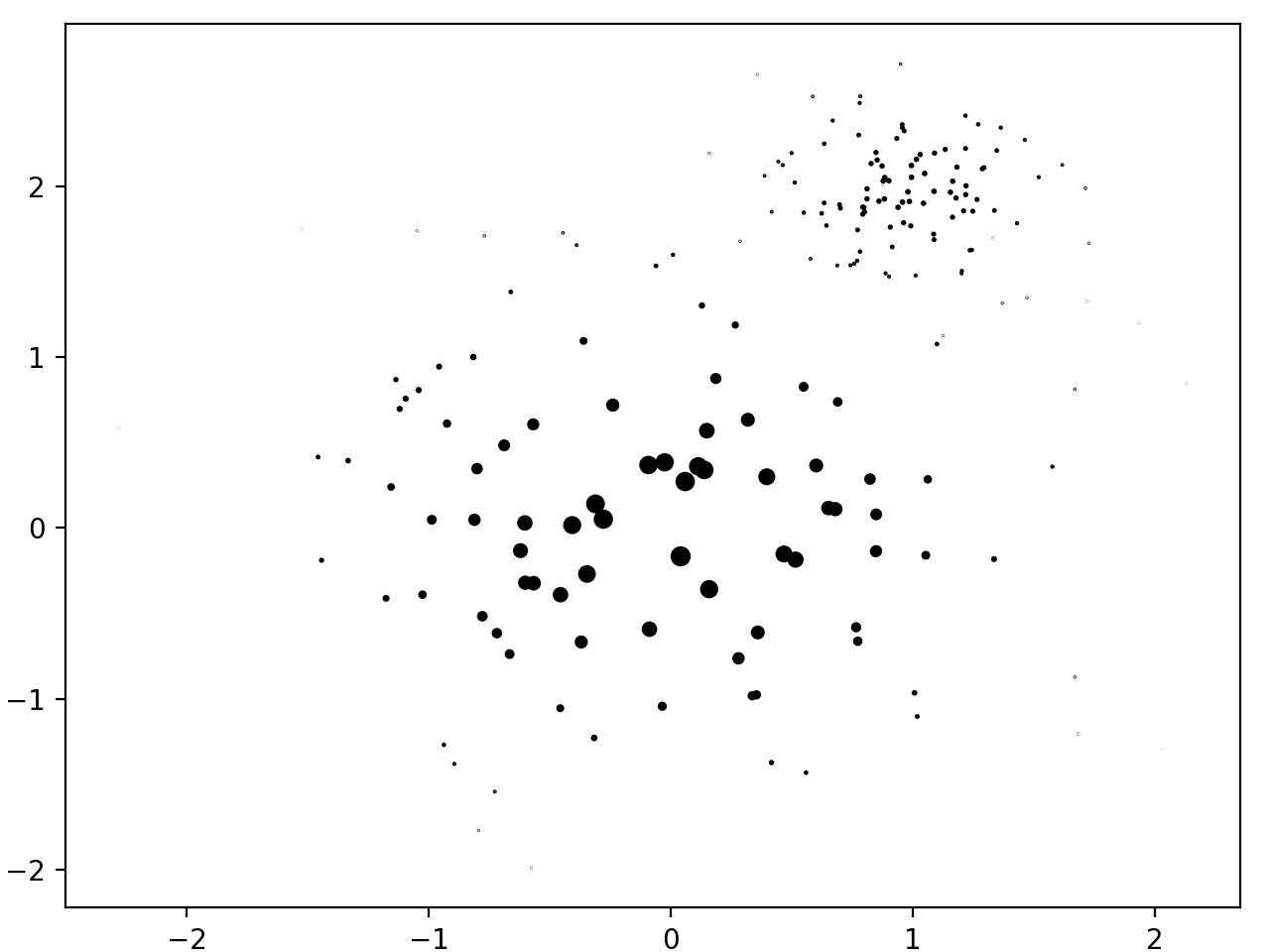}
        \end{tabular}
    \caption{An illustrative example of the routing Algorithm obtained by minimizing \ref{eq:loss} applied to a toy dataset, using cosine similarity. 
    The position of the bubbles corresponds to the lower level capsule votes in the higher level capsule space. 
    The votes were obtained by sampling two distinct multivariate Gaussian distributions with different means and covariances. 
    The bubble chart in (b), presents an overview of the toy dataset. The area of the bubbles, 
    encodes the activation probabilities of the corresponding capsule votes. 
    The activation probabilities were obtained by assigning the votes to one of the two distributions obtained after scaling each of the covariance matrices of the vote distributions. 
    For clarity, I normalized the activation probabilities, however, they need not be. 
    The bubble charts in (a), show the the optimal values of compatibility probabilities, 
    encoded by the area of the blue bubbles, and output capsule's pose,encoded by the yellow bubble, for different values of $\lambda_1$ and $\lambda_2$.   }
    \label{fig:scatters}
    \end{figure}

Figure \ref{fig:scatters}, contains an illustrative example, applied to a toy dataset,
of the derived clustering algorithm for different pairs of $\lambda_1$ and $\lambda_2$. 

Applied to the general routing procedure framework, presented in algorithm \ref{agreementgen}, 
we can take the \textit{compatibility} function to be equation \ref{eq:kernelcomp}.

\begin{equation} \label{eq:kernelcomp}
    compatibility_{\theta,\lambda_1,\lambda_2}(V, a_i, \mu,\gamma, s_i) \doteq (\frac{  a_i^{\frac{\lambda_2}{\lambda_1+\lambda_2}} exp( \frac{\langle \mu , v_i \rangle_{\theta}}{\lambda_1 + \lambda_2}) }{ \sum_{j \in \gamma}^{} a_j^{\frac{\lambda_2}{\lambda_1+\lambda_2}} exp( \frac{\langle \mu , v_j \rangle_{\theta}}{\lambda_1 + \lambda_2}) }, s_i )
\end{equation}

In this way, we obtain a \textit{compatibility} function that is parameterized by $\theta$, $\lambda_1$ and $\lambda_2$ which are learned discriminatively using back-propagation during the training process of the entire capsule network.
The remaining function for the definition of the routing procedure is the \textit{activation} which is present in \ref{eq:actikern}.

\begin{equation} \label{eq:actikern}
    activation_\beta(\mu, V, C, \gamma) \doteq sigmoid( \beta_1 \sum_{i \in \gamma}^{} c_i \langle \mu , v_i \rangle_{\theta} - \beta_2 D_{KL}(C\parallel A) + \beta_3 ), \beta_2 \geq 0, \beta_1 \geq 0
\end{equation}

In the context of convolutional routing, $\lambda_1$, $\lambda_2$, $\beta_1$, $\beta_2$, 
and $\beta_3$ are distinct for each channel in every convolutional routing layer. 
The parameters of the kernel function could either be shared across convolutional channel or layer, 
in the following sections, we assume they only shared across layers.

\subsection{Connectionist Approach}
\label{connectionistlabel}

An alternative approach to the more formal presented in the previous subsection, to modeling the routing procedure, 
is to allow the $activation$ and $compatibility$ functions in Algorithm \ref{agreementgen} to be Neural Networks. 
More precisely I employed a LSTM cell \cite{lstm}, which is designed to keep track of arbitrary long-term dependencies, 
in conjunction with two distinct neural networks to obtain a learnable routing mechanism that plays an active role throughout the iterations.

\begin{figure}
    \centering
    \includegraphics[scale=0.3]{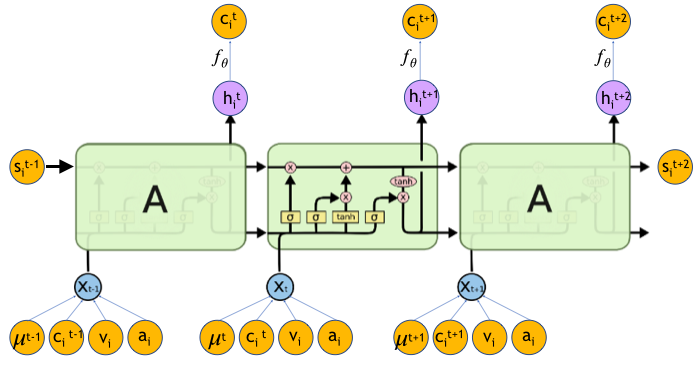}
    \setlength{\belowcaptionskip}{-32pt}
    \caption{Diagram of of the LSTM cell used in the routing mechanism(adapted from \cite{undlstm}).}
    \label{fig:lstmdig}
\end{figure}

Figure \ref{fig:lstmdig} presents a diagram of the LSTM cell employed. 
The input of the cell is a concatenation of the previous iteration intermediate output capsule pose $\mu$, 
the capsule vote from the $ith$ input capsule $v_i$ and the corresponding compatibility probabilities $c_i$ and the activation $a_i$. 
After initialized with zeros($s_0 = \mathbf{0} $), 
the cell state $s_i^t$ and hidden state $h_i^{t}$ are updated throughout the iterations to obtain representations that are subsequently fed to two distinct neural networks.

The neural network $f_\theta$ is applied to the outputs of the hidden state to produce the compatibility probabilities at the end of every iteration. 
Eventually, at the end of the iterative process, the cell states, correspondent to every input capsule, after combined, 
are fed to the neural network $g_\beta$ and result in the output capsule's final activation as indicated by \ref{eq:stateconcat}.
The definition of the compatibility function is presented in algorithm \ref{connectionistcomp}.

\begin{equation} \label{eq:stateconcat}
    activation_\beta(C, \gamma, S^t) \doteq g_\beta ( \sum_{i \in \gamma}^{} c_i s^{t}_{i} )
\end{equation}

\begin{algorithm}
\caption{Connectionist Approach} \label{connectionistcomp}
\begin{algorithmic}[1]
        \Procedure{$compatibility$}{$c_{i}, v_{i}, a_i, \mu, s_{i}^{t-1}$}
            \State $ x_{i}^{t} \gets concat(\mu, c_i, v_{i}, a_i)$
            \State $ h^{t}_{i}, s_{i}^{t} \gets LSTM(x_{i}^{t}, s_{i}^{t-1})$
            \State $c_{i} \gets f_\theta(h^{t}_{i})$
            \State \Return $c_{i}, s^{t}$
        \EndProcedure
    \end{algorithmic}
\end{algorithm}
The parameters from the LSTM and both neural networks are shared across every single input capsule. 
Additionally, the dimension of the LSTM's cell state and hidden state, as well as the architectures of the neural networks used, 
become hyperparameters.

\section{Experiments}
In order to evaluate the effectiveness of the proposed routing algorithms, when compared with the extension of \textbf{Expectation-Maximization}(EM) applied to a Gaussian Mixture Model, introduced in \cite{Hinton2018}, 
experiments were conducted in both MNIST dataset \cite{mnist} and smallNORB \cite{smallnorb}.
They consist in training the same Capsule Network architecture, for 100 epochs, with every routing procedure running for three iterations, for both of the proposed algorithms and the one present in \cite{Hinton2018} and comparing the respective test set results. 

The MNIST dataset was selected, in the early stages, as proof of concept. 
It is a large database of handwritten digits that has become the classic benchmark for machine learning models. 
The training set is composed of 60000 examples and the test set 10000.

\begin{figure}
    \centering
    \includegraphics[scale=0.4]{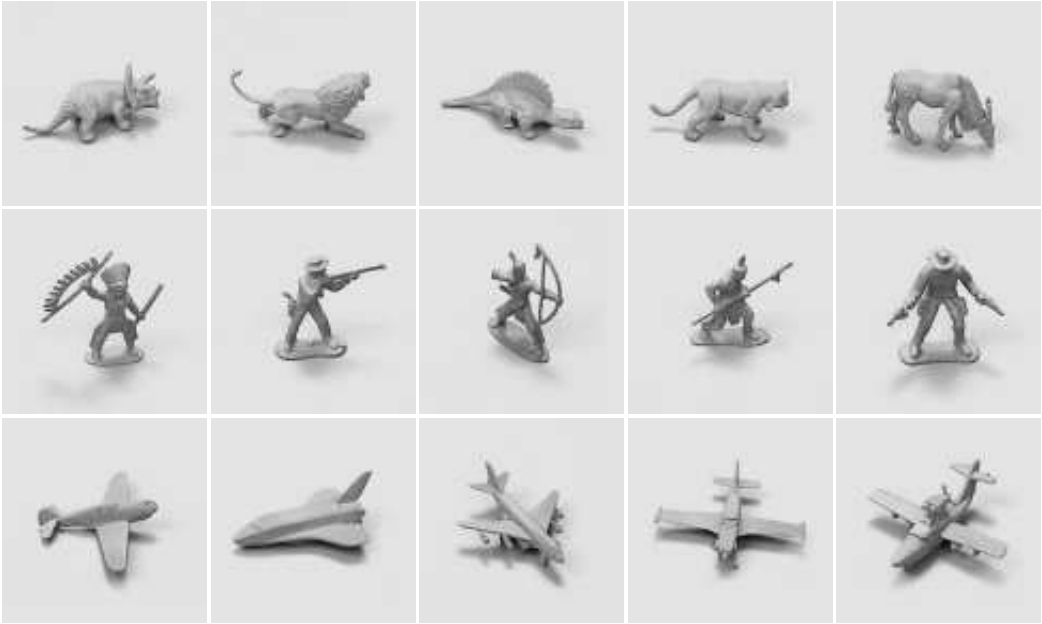}
    \caption{Examples of the classes animals, human figures and airplanes.}
    \label{fig:fgnorb}
\end{figure}

The smallNORB dataset was chosen due to it being much closer to natural images and yet devoid of context and color.
It is composed of 50 toys belonging to 5 generic categories(four-legged animals, human figures, airplanes, trucks, and cars).  
The objects were imaged by two cameras under 6 lighting conditions, 9 elevations (30 to 70 degrees every 5 degrees), 
and 18 azimuths (0 to 340 every 20 degrees). Figure \ref{fig:fgnorb} contain some sampled examples of the dataset.
The training set is composed of 5 instances of each category(24300 examples), and the test set of the remaining 5 instances(24300 examples).

For the experiments with smallNORB, this dataset was preprocessed precisely in the same manner as in \cite{Hinton2018}. 
The images were normalized and downsampled to $48 \times 48$. 
During training, they were randomly cropped into $32 \times 32$ patches and random brightness and contrast were added. 
During test and validation, however, each crop was correspondent to the center of the image.

All the experiments where made using a Tensorflow \cite{tensorflow2015-whitepaper} implementation of the underlying models \footnote{https://github.com/goncalorafaria/learning-inner-consensus}. 
The optimizer used was Adam \cite{Adam} with a learning rate of 3-e3 scheduled by an exponential decay.
Additionally, the models were trained using kaggle notebooks, approximately 40 hours each, a free research tool for machine learning that provides a Tesla P100 Nvidia GPU as part of an accelerated computing environment.

Apart from the implementation of the routing procedure, all of the models contained in the experiments instantiate the Capsule Network architecture present in Table \ref{table:architec} and use 4x4 matrices as poses. 
The model is a slight modification of the smaller Capsule Network used in \cite{Hinton2018}. 
More precisely, after the initial convolutions batch normalization \cite{batchnorm} is applied, and the transformation used is the one described in equation \ref{eq:transf}. 
Applied to smallNORB, the model has 86K parameters, excluding the ones pertaining to the different routing procedures.

\begin{table}[htb]
    %\caption{Capsule Network architecture}
    \centering
    \label{table:architec}
    \begin{tabular}{|lll|ll}
    \cline{1-3}
    Layer & Details  & Output shape   \\ \hline
    Input &  & $B \times 32 \times 32 \times 1$   \\
    Convolutional layer + relu + BatchNorm & K=5, S=2, Ch=64 & $B \times 16 \times 16 \times 64$ \\
    Primary Capsules  & K=1, S=1, Ch=8 & $B \times 16 \times 16 \times 8 \times (4 \times 4 + 1)$ \\
    Convolutional Capsule Layer 1 & K=3, S=2, Ch=16 & $B \times 7 \times 7 \times 16 \times (4 \times 4 + 1)$  \\
    Convolutional Capsule Layer 2  & K=3, S=1, Ch=16  & $B \times 5 \times 5 \times 16 \times ( 4 \times 4 + 1)$  \\
    Capule Class Layer & flatten, O=5 &  $B \times 1 \times 1 \times 5 \times (4 \times 4 + 1)$ \\ \hline
    \end{tabular}
    \medskip
    \setlength{\belowcaptionskip}{-35pt}
    \caption{ 
        Specification of the Capsule Network model used in the experiment with SmallNORB dataset. 
    \textbf{K} denotes convolutional kernel size, \textbf{S} stride, \textbf{Ch} number of output chanels, \textbf{O} number of classes and \textbf{B} the batch size.
    }
\end{table}

\begin{table}[htb]
    %\caption{ Connectionist approach routing parameters.}
    \centering
    \label{table:connect}
    \begin{tabular}{|llll|}
    \hline
 Layer & hidden layers in $f_\theta$ & hidden layers in $g_\beta$ & \begin{minipage}{40mm} ~\\ number of units in the lstm's hidden and cell states \\ \end{minipage} \\ \hline
    Convolutional Capsule Layer 1 &  [] & [] & 16 \\
    Convolutional Capsule Layer 2 &  [32,32] & [64,64] & 16 \\
    Capule Class Layer &   [64,64] & [124,124] & 16 \\ \hline
    \end{tabular}
    \medskip
    \setlength{\belowcaptionskip}{-10pt}
    \caption{
        The detailed specification of the hyperparameters used in the experiments with the Connectionist approach's routing mechanism. 
        The representation of the hidden layers is a list where the $ith$ list element corresponds to the number 
        of neurons in the $ith$ hidden layer.
    }
\end{table}

The model present in table \ref{table:architec} contains three layers to which routing is applied.
When used in the described experiments, the routing algorithm presented in section \ref{connectionistlabel}, used the hyperparameters described in table \ref{table:connect}. 
Moreover, the routing algorithm presented in section \ref{klearning} used the kernel defined in equation \ref{eq:kernelused}, a combination of gaussian kernels\cite{Risser2011SimultaneousMR}. The first two layers in the Capsule Network have Q = 4 and the last one has Q = 10.

\begin{equation} \label{eq:kernelused}
    \langle x , x' \rangle_{\theta} \doteq \sum^{Q}_{q=1} \theta^q_{1} exp( - \frac{\| x - x' \|^2}{ 2(\theta^q_{2})^2 } )
\end{equation}
The choice of routing hyperparameters was based on making the routing procedure in the deeper layers have more parameters. 
What motivated this design choice was the intuition that, the more complex the features, further complex needed to be the routing.
Ideally, if there were not computational limitations, these parameters would have been chosen using a randomized grid search with cross-validation or other more suitable hyperparameter search method.

\begin{table}[]
    \centering
    \label{table:error}
    \begin{tabular}{|lll|}
    \hline
     Routing & MNIST & smallNORB  \\ \hline
     Connectionist &  0.53\% & 6.7\%  \\
     EM \cite{ibm2019} & (not available) & 6.3\% \\
     EM (www0wwwjs1)\footnote{https://github.com/www0wwwjs1/Matrix-Capsules-EM-Tensorflow} & 0.9\% & 8.2\% \\
     EM (original authors)\cite{Hinton2018} & (not available) &  2.2\% \\
     Similarity Learning & 1.0\% & 8.19\%  \\ \hline
    \end{tabular}
    \medskip
    \setlength{\belowcaptionskip}{-5pt}
    \caption{Test set percentual error rate obtained in the experiments. 
    During the optimization process, starting from the tenth epoch, the parameters for each model, were recorded(5000 instances). 
    The reported results correspond to the test set percentual error of the parameters that achieved higher accuracy on the validation set(10\% original training set).
    Given that the original authors of EM routing did not make their implementations publicly available, 
    I also compare my work with the two best open-source versions that I have found.}
\end{table}

The results pertaining to all of the experiments are contained in table \ref{table:error}.
The table contains the percentual error rate achieved in the evaluation of the models.
\section{Discussion}

The results we obtained on smallNORB are far from the ones reported by \cite{Hinton2018}.
However, when in comparison with the best open-source versions, the same does not hold. 
I speculate that the original authors did not present all of the needed implementation details, 
which would indeed explain the observed discrepancy.
When in comparision with the best open-source versions, 
the Connectionist approach is on par or surpasses them while the Similarity Learning approach does not.

The main challenge I encountered while fitting the models was the computational time required. 
Although the convolutional Capsule networks require less floating-point operations and fewer training parameters, 
than any comparable CNN, they are much slower and ran out of memory with relatively small models. 
This is due to Tensorflow beeing optimized for CNNs, 
and not having natively the operations required to compute the Capsule's routing convolutions. 
To compute the routing Convolutions, for each native TensorFlow operation employed, 
TensorFlow's functional API makes a copy of the underlying tensors, which greatly increases the memory requirements. 

\section{Conclusion}
In this paper, 
I have presented two distinct ways to learn Capsule Network's inner routing mechanism using backpropagation.
The Similarity Learning and the Connectionist approach.
The proposed methods build on the work of \cite{Sabour2017} and \cite{Hinton2018} 
by augmenting the part whole relationships that can be modeled.

The experimental results show that the Connectionist approach is on par or surpasses
with the best open-source implementations of Capsule Networks using EM while the Similarity Learning does not.
When in comparison with the original paper, with EM routing, the results of the introduced approaches fall short.
I suspect this is due to missing implementation details that are not mentioned in the original paper. 
When these are made available, 
I expect the results from the routing mechanisms proposed to considerably improve.

Future work will focus on additional applications, of the proposed extension of capsule networks, 
in different domains and computer vision tasks, 
experiment with distinct kernel parameters for distinct convolutional channels, 
and study the interpretability of the proposed models.
Particularly, for applications in object detection tasks, 
I intend to use the capsule poses of the last layer to more effectively identify the location of bounding boxes.
The study of the interpretability of the proposed models will be
aimed at facilitating model building and present mechanisms for verifying the usefulness of the model predictions. 
Aditionally, I intend to augment the Similarity learning approach to more general divergence functions such as the Jensen–Tsallis divergence.

\bigskip

\textbf{ACKNOWLEDGMENTS} A special thank you to the Gulbenkian Foundation, the scientific committee of the Artificial Intelligence Program and this project's tutor Professor Cesar Analide. 
Addittionally, I would also like to thank the reviewers for their
constructive comments which helped to improve the proposed methodology.

\bibliographystyle{splncs}
\bibliography{ref}

\begin{thebibliography}{10}

\bibitem{Rosenblatt1958}
Rosenblatt, F.:
\newblock {The perceptron: A probabilistic model for information storage and
  organization in the brain}.
\newblock Psychological Review (1958)

\bibitem{LeCun1989}
LeCun, Y.:
\newblock {Handwritten Digit Recognition with a Back-Propagation Network}.
\newblock Advances in Neural Information Processing Systems (1989)

\bibitem{Krizhevsky2012}
Krizhevsky, A., Sutskever, I., Hinton, G.E.:
\newblock {ImageNet Classification with Deep Convolutional Neural Networks}.
\newblock In: ImageNet Classification with Deep Convolutional Neural Networks.
  (2012)

\bibitem{HintonG.DengL.YuD.DahlG.MohamedA.JaitlyN.Kingsbury2012}
{Hinton, G., Deng, L., Yu, D., Dahl, G., Mohamed, A., Jaitly, N., {\ldots}
  Kingsbury}, B.:
\newblock {Deep Neural Networks for Acoustic Modeling in Speech Recognition.}
\newblock IEEE Signal Processing Magazine (2012)

\bibitem{Bengio2003}
Bengio, Y., Ducharme, R., Vincent, P., Jauvin, C.:
\newblock {A Neural Probabilistic Language Model}.
\newblock In: Journal of Machine Learning Research. (2003)

\bibitem{Google}
Yonghui~Wu, Mike~Schuster, Z.C.Q.V.L.M.N.W.M.M.K.Y.C.Q.G.K.M.
\newblock (2016)

\bibitem{Kleesiek2016}
Kleesiek, J., Urban, G., Hubert, A., Schwarz, D., Maier-Hein, K., Bendszus, M.,
  Biller, A.:
\newblock {Deep MRI brain extraction: A 3D convolutional neural network for
  skull stripping}.
\newblock NeuroImage (2016)

\bibitem{Dou2016}
Dou, Q., Chen, H., Yu, L., Zhao, L., Qin, J., Wang, D., Mok, V.C., Shi, L.,
  Heng, P.A.:
\newblock {Automatic Detection of Cerebral Microbleeds from MR Images via 3D
  Convolutional Neural Networks}.
\newblock IEEE Transactions on Medical Imaging (2016)

\bibitem{Sermanet2011}
Sermanet, P., Lecun, Y.:
\newblock {Traffic sign recognition with multi-scale convolutional networks}.
\newblock In: Proceedings of the International Joint Conference on Neural
  Networks. (2011)

\bibitem{Szegedy2015}
Szegedy, C., Liu, W., Jia, Y., Sermanet, P., Reed, S., Anguelov, D., Erhan, D.,
  Vanhoucke, V., Rabinovich, A.:
\newblock {Going deeper with convolutions}.
\newblock In: Proceedings of the IEEE Computer Society Conference on Computer
  Vision and Pattern Recognition. (2015)

\bibitem{Hinton2011}
Hinton, G.E., Krizhevsky, A., Wang, S.D.:
\newblock {Transforming auto-encoders}.
\newblock In: Lecture Notes in Computer Science (including subseries Lecture
  Notes in Artificial Intelligence and Lecture Notes in Bioinformatics). (2011)

\bibitem{Sabour2017}
Sara~Sabour, N.F., Hinton, G.E.:
\newblock {Dynamic routing between capsules}.
\newblock In: Advances in Neural Information Processing Systems. (2017)

\bibitem{Hinton2018}
Hinton, G., Sabour, S., Frosst, N.:
\newblock {Matrix capsules with EM routing}.
\newblock In: International Conference on Learning Representations. (2018)

\bibitem{Wang2018}
Wang, D., Lui, Q.:
\newblock {An Optimization View on Dynamic Routing Between Capsules}.
\newblock Journal of Geotechnical and Geoenvironmental Engineering (2018)

\bibitem{av1}
Deli{\`{e}}ge, A., Cioppa, A., Droogenbroeck, M.V.:
\newblock Hitnet: a neural network with capsules embedded in a hit-or-miss
  layer, extended with hybrid data augmentation and ghost capsules.
\newblock CoRR \textbf{abs/1806.06519} (2018)

\bibitem{av2}
{Xiang}, C., {Zhang}, L., {Tang}, Y., {Zou}, W., {Xu}, C.:
\newblock Ms-capsnet: A novel multi-scale capsule network.
\newblock IEEE Signal Processing Letters \textbf{25} (2018)  1850--1854

\bibitem{av3}
Amer, M., Maul, T.:
\newblock Path capsule networks.
\newblock CoRR \textbf{abs/1902.03760} (2019)

\bibitem{av4}
{O' Neill}, J.:
\newblock {Siamese Capsule Networks}.
\newblock arXiv e-prints (2018)  arXiv:1805.07242

\bibitem{av5}
Rawlinson, D., Ahmed, A., Kowadlo, G.:
\newblock Sparse unsupervised capsules generalize better.
\newblock CoRR \textbf{abs/1804.06094} (2018)

\bibitem{av6}
Ma, D., Wu, X.:
\newblock Tcdcaps: Visual tracking via cascaded dense capsules.
\newblock CoRR \textbf{abs/1902.10054} (2019)

\bibitem{av7}
do~Rosario, V.M., Borin, E., Jr., M.B.:
\newblock The multi-lane capsule network {(MLCN)}.
\newblock CoRR \textbf{abs/1902.08431} (2019)

\bibitem{av8}
Zhao, Y., Birdal, T., Deng, H., Tombari, F.:
\newblock 3d point-capsule networks.
\newblock CoRR \textbf{abs/1812.10775} (2018)

\bibitem{Vaswani2017AttentionIA}
Vaswani, A., Shazeer, N., Parmar, N., Uszkoreit, J., Jones, L., Gomez, A.N.,
  Kaiser, L., Polosukhin, I.:
\newblock Attention is all you need.
\newblock In: NIPS. (2017)

\bibitem{ibm2019}
Gritzman, A.D.
\newblock (2019)

\bibitem{Li2008}
Li, M.J., Ng, M.K., Cheung, Y.M., Huang, J.Z.:
\newblock {Agglomerative fuzzy K-Means clustering algorithm with selection of
  number of clusters}.
\newblock IEEE Transactions on Knowledge and Data Engineering (2008)

\bibitem{kldv}
Kullback, S., Leibler, R.A.:
\newblock On information and sufficiency.
\newblock The Annals of Mathematical Statistics \textbf{22} (1951)  79--86

\bibitem{lstm}
Hochreiter, S., Schmidhuber, J.:
\newblock Long short-term memory.
\newblock Neural computation \textbf{9} (1997)  1735--80

\bibitem{undlstm}
:
\newblock Understanding lstm networks.
\newblock (\url{http://colah.github.io/posts/2015-08-Understanding-LSTMs/})
  Accessed: 2019-08-15.

\bibitem{mnist}
LeCun, Y., Cortes, C.:
\newblock The mnist database of handwritten digits.
\newblock (2005)

\bibitem{smallnorb}
LeCun, Y., Huang, F.J., Bottou, L.:
\newblock Learning methods for generic object recognition with invariance to
  pose and lighting.
\newblock Proceedings of the 2004 IEEE Computer Society Conference on Computer
  Vision and Pattern Recognition, 2004. CVPR 2004. \textbf{2} (2004)  II--104
  Vol.2

\bibitem{tensorflow2015-whitepaper}
Abadi, M., Agarwal, A., Barham, P., Brevdo, E., Chen, Z., Citro, C., Corrado,
  G.S., Davis, A., Dean, J., Devin, M., Ghemawat, S., Goodfellow, I., Harp, A.,
  Irving, G., Isard, M., Jia, Y., Jozefowicz, R., Kaiser, L., Kudlur, M.,
  Levenberg, J., Man\'{e}, D., Monga, R., Moore, S., Murray, D., Olah, C.,
  Schuster, M., Shlens, J., Steiner, B., Sutskever, I., Talwar, K., Tucker, P.,
  Vanhoucke, V., Vasudevan, V., Vi\'{e}gas, F., Vinyals, O., Warden, P.,
  Wattenberg, M., Wicke, M., Yu, Y., Zheng, X.:
\newblock {TensorFlow}: Large-scale machine learning on heterogeneous systems
  (2015) Software available from tensorflow.org.

\bibitem{Adam}
Kingma, D.P., Ba, J.:
\newblock Adam: A method for stochastic optimization.
\newblock CoRR \textbf{abs/1412.6980} (2014)

\bibitem{batchnorm}
Ioffe, S., Szegedy, C.:
\newblock Batch normalization: Accelerating deep network training by reducing
  internal covariate shift.
\newblock CoRR \textbf{abs/1502.03167} (2015)

\bibitem{Risser2011SimultaneousMR}
Risser, L., Vialard, F.X., Wolz, R., Murgasova, M., Holm, D.D., Rueckert, D.:
\newblock Simultaneous multi-scale registration using large deformation
  diffeomorphic metric mapping.
\newblock IEEE Transactions on Medical Imaging \textbf{30} (2011)  1746--1759

\end{thebibliography}

\end{document}